\documentclass[10pt,Afour,sagev,times]{sagej}
\usepackage{lmodern}
\usepackage[latin9]{inputenc}
\usepackage{verbatim}
\usepackage{prettyref}
\usepackage{textcomp}
\usepackage{graphicx}
\usepackage[unicode=true,
 bookmarks=false,
 breaklinks=false,pdfborder={0 0 1},backref=section,colorlinks=false]
 {hyperref}
\hypersetup{
 colorlinks,bookmarksopen,bookmarksnumbered,citecolor=red,urlcolor=red}

\makeatletter
\usepackage{moreverb}
\usepackage{url}
\PassOptionsToPackage {hyphens}{url} 

\newcommand{\BibTeX}{{\rmfamily B\kern-.05em \textsc{i\kern-.025em b}\kern-.08em
T\kern-.1667em\lower.7ex\hbox{E}\kern-.125emX}}

\def\volumeyear{2016}

\makeatother

\begin{document}
\runninghead{Teikari et al.}
\title{Embedded deep learning in ophthalmology: \textit{Making ophthalmic
imaging smarter}}
\author{Petteri Teikari\affilnum{1,2}, Raymond P. Najjar\affilnum{1,3},
Leopold Schmetterer \affilnum{1,2,4,5} and Dan Milea\affilnum{1,3,6}}

\corrauth{Petteri Teikari, Visual Neurosciences group, Singapore
Eye Research Institute, Singapore. Academia, 20 College Road, Discovery
Tower Level 6, Singapore 169856}

\email{petteri.teikari@gmail.com}

\affiliation{\affilnum{1} Visual Neurosciences Department, Singapore Eye Research Institute, Singapore\\
\affilnum{2} Advanced Ocular Imaging, Lee Kong Chian School of Medicine, Nanyang Technological University, Singapore\\
\affilnum{3} Ophthalmology and Visual Sciences Academic Clinical Program, Duke-NUS Medical School, National University of Singapore, Singapore\\
\affilnum{4} Center for Medical Physics and Biomedical Engineering, Medical University of Vienna, Austria\\
\affilnum{5} Christian Doppler Laboratory for Ocular and Dermal Effects of Thiomers, Medical University of Vienna, Austria\\
\affilnum{6} Neuro-Ophthalmology Department, Singapore National Eye Centre, Singapore}

\begin{abstract}

Deep learning has recently gained high interest in ophthalmology,
due to its ability to detect clinically significant features for diagnosis
and prognosis. Despite these significant advances, little is known
about the ability of various deep learning systems to be embedded
within ophthalmic imaging devices, allowing automated image acquisition.
In this work, we will review the existing and future directions for
``active acquisition'' embedded deep learning,
leading to as high quality images with little intervention by the
human operator. In clinical practice, the improved image quality should
translate into more robust deep learning-based clinical diagnostics.
Embedded deep learning will be enabled by the constantly improving
hardware performance with low cost. We will briefly review possible
computation methods in larger clinical systems. Briefly, they can
be included in a three-layer framework composed of edge, fog and cloud
layers, the former being performed at a device-level. Improved egde
layer performance via ``active acquisition''
serves as an automatic data curation operator translating to better
quality data in electronic health records (EHRs), as well as on the
cloud layer, for improved deep learning-based clinical data mining. 

\end{abstract}


\maketitle

\section{Introduction}

Recent years have seen an explosion in the use of deep learning algorithms
for medical imaging \cite{litjens2017asurvey,hinton2018deeplearningtextemdasha,ching2018opportunities},
including ophthalmology \cite{schmidt-erfurth2018artificial,ting2018aifor,hogartycurrent,lee2017machine,fauw2018clinically2}.
Deep learning has been very efficient in detecting clinically significant
features for ophthalmic diagnosis\cite{ting2017development,fauw2018clinically2}
and prognosis\cite{schmidt-erfurth2018machine,wen2018forecasting}.
Recently, Google Brain demonstrated how one can, surprisingly, predict
subject's cardiovascular risk, age and gender from a fundus image
\cite{poplin2018prediction}, a task impossible for an expert clinician.

Research effort has so far focused on the development of post\textendash hoc
deep learning algorithms for already acquired datasets \cite{ting2017development,fauw2018clinically2}.
There is, however, growing interest for embedding deep learning at
the medical device level itself for real-time image quality optimization,
with little or no operator expertise. Most of the clinically available
fundus cameras and optical coherence tomography (OCT) devices require
the involvement of a skilled operator in order to achieve satisfactory
image quality, for clinical diagnosis. Ophthalmic images display inherent
quality variability due to both technical limitations of the imaging
devices, and individual ocular characteristics. Recent studies in
hospital settings have shown that 38\% of nonmydriatic fundus images
for diabetic screening \cite{rani2018analysis}, and 42-43\% of spectral
domain (SD)-OCTs acquired for patients with multiple sclerosis\cite{tewarie2012theoscarib}
did not have acceptable image quality for clinical evaluation. 

Desktop retinal cameras have been increasingly replaced by portable
fundus cameras in standalone format \cite{roesch2017automated,monroy2017clinical,chopra2017humanfactor}
or as smartphone add-ons \cite{kim2018asmartphonebased}, making the
retinal imaging less expensive and accessible to various populations.
The main drawback of the current generation portable fundus camera
is the lower image quality. Some imaging manufacturers have started
to include image quality assessment algorithms to provide a feedback
for the operator to either re-acquire the image or accept it \cite{katuwal2018automated}.
To the best of our knowledge, no current commercial system is automatically
reconstructing ``the best possible image'' from multiframe image
acquisitions.

Embedding of more advanced algorithms and high computation power at
the camera level can be referred to  as ``smart camera architectures''
\cite{brea2018special}, with or without the use of deep learning.
For example, Google launched its Clips camera, and Amazon Web Services
(AWS) its DeepLens camera which are capable of running deep learning
models within the camera itself without relying on external processing
Verily, the life sciences research organization of Alphabet Inc,
partnered with Nikon and Optos to integrate deep learning algorithms
for fundus imaging and diabetic retinopathy screening\footnote{\href{https://verily.com/projects/interventions/retinal-imaging/}{https://verily.com/projects/interventions/retinal-imaging/}}.
Similar implementation of ``intelligence'' at the device-level is
happening in various other medical fields \cite{zhang2018influence},
including portable medical ultrasound imaging, with more of the traditional
signal processing being accelerated graphics processing units (GPUs)
\cite{gobl2018supraopen}, with the deep learning integrated at the
device level \cite{jarosik2018waveflow}.

There are various ways of distributing the signal processing from
data acquisition to clinical diagnostics. For example, the use of
fundus cameras in remote locations with no internet access requires
all the computations to be performed within the device itself, a system
which has been implemented by SocialEyes, for retinal screening on
GPU-accelerated tablets\cite{hansen2016socialeyes}. This computing
paradigm, known as \emph{edge computing} \cite{shi2016edgecomputing},
is based on locally performed computations, on the ``edge'' \cite{cuff2018getting,harris2018thenext},
as opposed to cloud computing in which the fundus image is transmitted
over the internet to a remote\emph{ cloud} GPU server, allowing subsequent
image classification. In some situations, when there is a need for
multi-layer computational load distribution, additional nodes are
inserted between the edge device and the cloud, a computation paradigm
known as \emph{mist }\cite{barik2018mistdata}\emph{ }or \emph{fog
computing}\cite{xu2018quantitative}. This situation applies typically
to Internet-of-Things (IoT) medical sensors, which often have very
little computational capability \cite{farahani2018towards}.

\begin{figure*}
\textbf{\includegraphics[width=2\columnwidth]{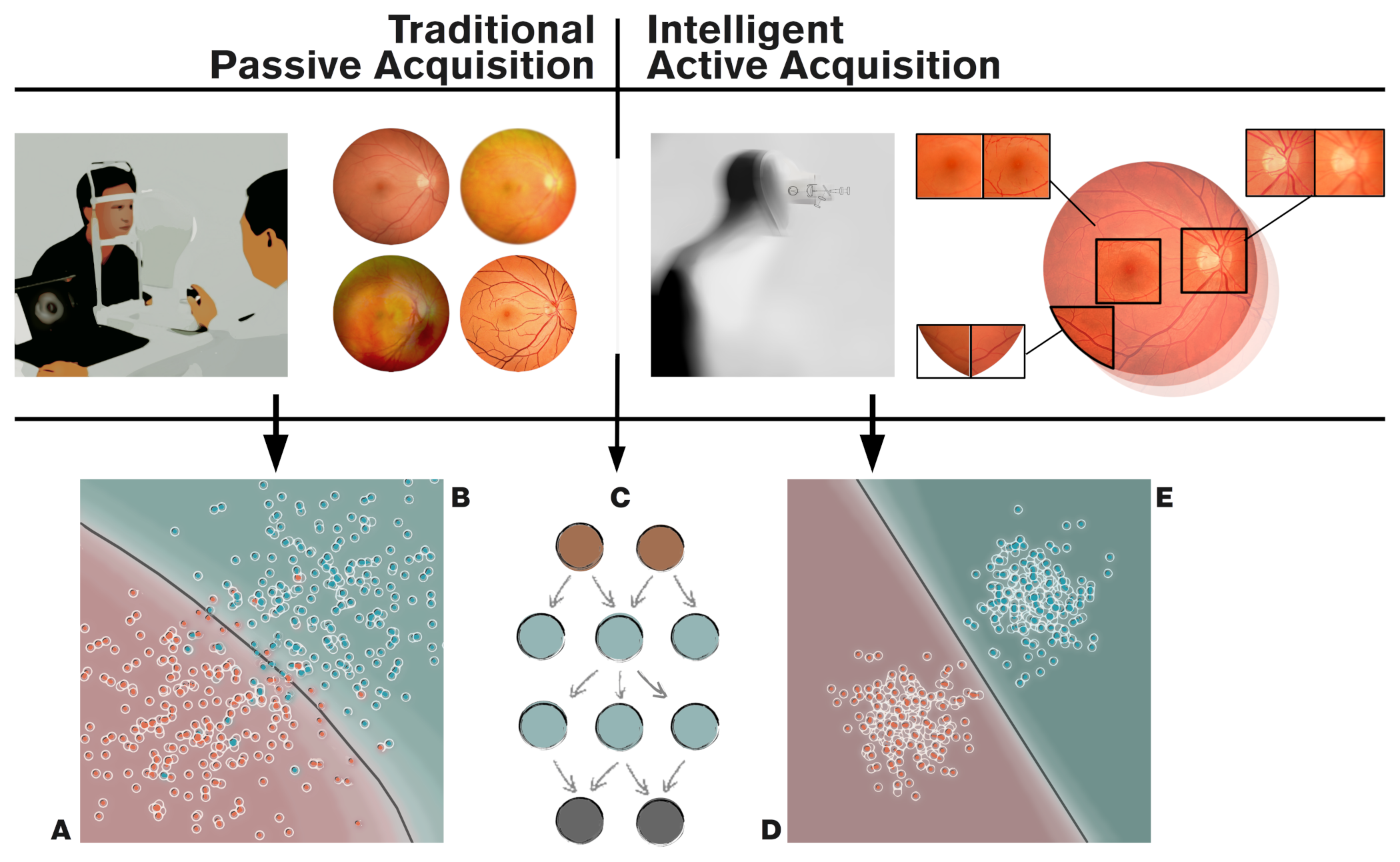}}

\caption{Comparison between traditional passive acquisition and intelligent
active acquisition approaches for fundus imaging. \textbf{(top-left)}
In passive acquisition, the healthcare professional manually aligns
the camera and decides the best moment for image acquisition. This
acquisition has to be often repeated, especially if the patient is
not compliant, if the pupils are not dilated, or if there are media
opacities, i.e. cornea scar, cataract, etc. \textbf{(top-right)} In
an ``intelligent'' active acquisition process, the device is able
vary imaging parameters, and iterates automatically frames until the
deep learning is been able to reconstruct an image of satisfactory
quality. \textbf{(bottom)} This intelligent acquisition serves as
automated data curation operator for diagnostic deep learning networks
(C)\cite{ting2017development,fauw2018clinically2} leading to improved
deep leading to better class separation (healthy D vs. disease E).
In traditional passive acquisition, the image quality is less consistent
leading to many false positives {[}patient from disease population
B (cyan) is classified as healthy A (red){]} and negatives {[}patient
from healthy population A (red) is classified as disease B (cyan){]}.
The gray line represents the decision boundary of the classifier \cite{fawzi2017classification},
and each point represent one patient. \label{fig:Intro-figure}}
\end{figure*}

The main aim of the current review is to summarize the current knowledge
related to device-level (\emph{edge }computing) deep learning. We
will refer to this as ``active acquisition'', for improved ophthalmic
diagnosis via optimization of image quality\textbf{ }(\prettyref{fig:Intro-figure}).
We will also overview various possibilities of computing platforms
integrate into the typical clinical workflow with a focus on standard
retinal imaging techniques (i.e. fundus photography and OCT).

\section{Embedded ophthalmic devices}

\subsection{Emerging intelligent retinal imaging}

The increased prevalence of ophthalmic conditions affecting the retinas
and optic nerves of vulnerable populations prompts higher access to
ophthalmic care both in developed \cite{lee2017disparities} and developing
countries \cite{sommer2014challenges}. This translates into an increased
need of more efficient screening, diagnosis and disease management
technology, operated with no or little training both in clinical settings,
or even at home \cite{roesch2017automated}. Although paraprofessionals
with technical training are currently able to acquire fundus images,
a third of these images may not be of satisfactory quality, being
non-gradable \cite{davila2017predictors}, due to reduced transparency
of the ocular media.

Acquisition of such images may be even more difficult in non-ophthalmic
settings, such as Emergency Departments \cite{hassen2018alleye}.
Recent attempts have aimed to automate retinal imaging processing
using a clinical robotic platform InTouch Lite (InTouch Technologies,
Inc., Santa Barbara, CA, USA) \cite{martel2015comparative}, or by
integrating a motor to the fundus camera for automated pupil tracking
(Nexy, Next Sight, Prodenone, Italy) \cite{2018nexyrobotic}. These
approaches have not been validated clinically, and are based on relatively
slow motors, possibly not adapted to clinically challenging situations.
Automated acquisition becomes even more important with the recent
surge of many smartphone-based fundus imagers \cite{barikian2018smartphone}.
Due to the pervasiveness of smartphones, this approach would represent
a perfect tool for non-eye specialists \cite{bifolck2018smartphone}. 

Similarly to fundus imaging, OCT systems are getting more portable
and inexpensive and would benefit from easier and robust image acquisition
\cite{chopra2017humanfactor,kim2018designand,monroy2017clinical}.
Kim \emph{et al. }\cite{kim2018designand} developed a low-cost experimental
OCT system at a cost of US\$ 7,200 using a microelectromechanical
system (MEMS) mirror \cite{lin2015progress} with a tunable variable
focus liquid lens to simplify the design of scanning optics, with
inexpensive Arduino Uno microcontroller \cite{teikari2012aninexpensive}
and GPU-accelerated mini PC handling the image processing. The increased
computing power from GPUs enables some of the hardware design compromises
to be offset through computational techniques\cite{altmann2018quantuminspired,liu2017computational}.
For example Tang \emph{et al. }\cite{tang2016gpubased} employed
three GPU units for real-time computational adaptive optics system,
and recently Maloca \emph{et al. }\cite{maloca2018highperformance}
employed GPUs for volumetric OCT in virtual reality environment for
enhanced visualization in medical education.

\subsection{Active Data Acquisition}

The computationally heavier algorithms made possible by the increased
hardware performance can be roughly divided into two categories: 1)
``passive'' single-frame processing, and 2) ``active'' multi-frame
processing . In our nomenclature, the ``passive'' techniques refer
to the standard way of acquiring ophthalmic images in which an operator
takes an image, which is subsequently subjected to various image enhancement
algorithms before being analyzed either by clinician or graded automatically
by an algorithm \cite{abr`amoff2018pivotal}. In ``active'' image
acquisition, multiple frames of the same structure are obtained with
either automatic reconstruction, or with interactive operator-assisted
reconstruction of the image. In this review, we will focus on the
``active'' paradigm, where clinically meaningful images would be
reconstructed automatically from multiple acquisitions with varying
image quality. 

One example for the active acquisition in retinal imaging is the 'Lucky
imaging' approach \cite{samaniego2014mobilevision,lawson2016methods},
in which multiple frames are acquired in quick succession assuming
that at least some of the frames are of good quality. In magnetic
resonance imaging (MRI), a 'prospective gating scheme' is proposed
for acquiring because motion-free image acquisition is possible between
the cardiovascular and respiration artifacts, iterating the imaging
until satisfactory result is achieved \cite{kinchesh2018prospective}.
For three-dimensional 3D Computed Tomography (CT), an active reinforcement
learning based algorithm was used to detect missing anatomical structures
from incomplete volume data \cite{ghesu2018towards}, and trying to
re-acquire the missing parts instead of relying just on post-acquisition
inpainting \cite{skalicligvoxel}. In other words, the active acquisition
paradigms have some level of knowledge of acquisition completeness
or uncertainty based on ideal images for example via ``active learning''
framework \cite{gal2017deepbayesian}, or via recently proposed Generative
Query Networks (GQN) \cite{eslami2018neuralscene}.

To implement active data acquisition on an ophthalmic imaging device,
we need to define a\emph{ loss function} (error term for the deep
learning network to minimize) to quantify the ``goodness'' of the
image either directly from the image, or using some auxiliary sensors
and actuators, to drive the automatic reconstruction process. For
example, eye movement artifacts during acquisition of OCT can significantly
degrade the image quality \cite{baghaie2017involuntary}, and we would
like to quantify the retinal motion either from the acquired frames
itself \cite{sheehy2012highspeed}, or by using auxiliary sensors
such as digital micromirror device (DMD) \cite{vienola2018invivo}.
The latter approach has also been applied for correction of light
scatter by opaque media\cite{turpin2018lightscattering}. Due to the
scanning nature of OCT, one can re-acquire the same retinal volume,
and merge only the subvolumes that were sampled without artifacts
\cite{carrasco-zevallos2016pupiltracking,chen2018eyemotioncorrected}.

\subsection{Deep learning-based retinal image processing}

Traditional single-frame OCT signal processing pipelines have employed
GPUs allowing real-time signal processing \cite{zhang2010realtime,wieser2014highdefinition}.
GPUs have been increasingly in medical image processing even before
the recent popularity of deep learning \cite{eklund2013medical}.
The GPUs are becoming essentially obligatory with contemporary high
speed OCT systems\cite{klein2017highspeed}. The traditional image
restoration pipelines employ the intrinsic characteristics of the
image in tasks such as denoising \cite{li2017statistical}, and deblurring
\cite{liu2009deconvolution} without considering image statistics
of a larger dataset. 

Traditionally these multi-frame reconstruction algorithms have been
applied after the acquisition without real-time consideration of the
image quality of the individual frames. Retinal multi-frame acquisition
such as fundus videography can exploit the redundant information across
the consecutive frames, and improve the image degradation model over
single-frame acquisition \cite{bian2013multiframe,devalla2018adeep}.
Köhler et al. \cite{kohler2014multiframe} demonstrated how a multi-frame
super-resolution framework can be used to reconstruct a single high-resolution
image from sequential low-resolution video frames. Stankiewicz \emph{et
al. \cite{stankiewicz2016matching} }implemented a similar framework
for reconstructing super-resolved volumetric OCT stacks from several
low quality volumetric OCT scans. Neither of these approaches, however,
applied the reconstruction in real-time. 

\begin{figure*}
\textbf{\includegraphics[width=2\columnwidth]{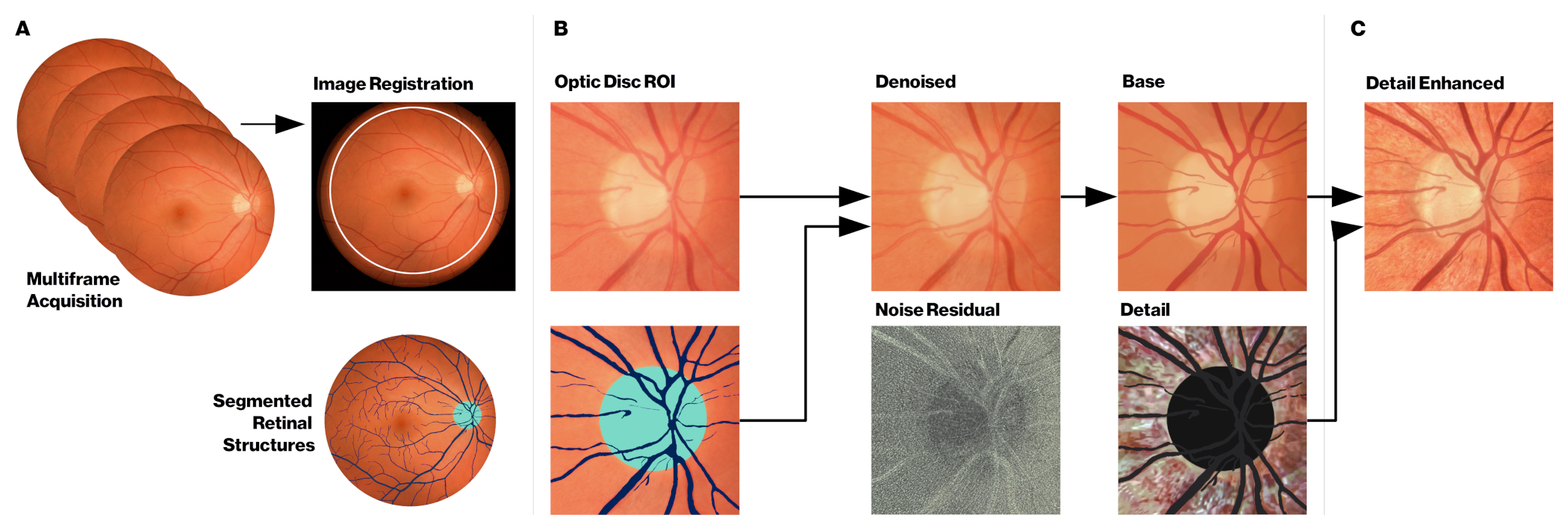}}

\caption{Typical image processing operators used in retinal image processing
that are illustrated with 2D fundus images for simplicity. \textbf{(A)}
Multiple frames are acquired in a quick succession, which are then
registered (aligned) with semantic segmentation for clinically meaningful
structures such as vasculature (in blue) and optic disc (in green).
\textbf{(B) }Region-of-interest (ROI) zoom on optic disc of the registered
image. The image is denoised with shape priors from the semantic segmentation
to help the denoising to keep sharp edges. The noise residual is normalized
for visualization showing some removal of structural information.
The denoised image is decomposed \cite{xu2011imagesmoothing} into
\emph{base }that contain the texture-free structure (edge-aware smoothing),
and the \emph{detail} that contains the residual texture without the
vasculature and optic disc. \textbf{(C) }An example of how the decomposed
parts can be edited ``layer-wise'' \cite{innamorati2017decomposing}
and combined to \emph{detail enhanced }image, in order to allow for
optimized visualization of the features of interest. \label{fig:imageProc-operators}}
\end{figure*}

In practice, all of the traditional image processing algorithms can
be updated for deep learning framework (\prettyref{fig:imageProc-operators}).
The ``passive'' approaches using input-output pairs to learn image
processing operators range from updating individual processing blocks
\cite{balakrishnan2018anunsupervised}, to joint optimization of multiple
processing blocks \cite{diamond2017dirtypixels,liu2017whenimage},
or training an end-to-end network such as DeepISP (ISP, Image Signal
Processor) to handle image pipeline from raw image towards the final
edited image \cite{schwartz2018deepisp}. The DeepISP network was
developed as offline algorithm \cite{schwartz2018deepisp}, with no
real-time optimization of camera parameters during acquisition. Sitzmann
\emph{et al. }\cite{sitzmann2018endtoend} extended the idea even
further by jointly optimizing the imaging optics and the image processing
for extended depth-of-field and super-resolution.

With deep learning, many deep image restoration networks have been
proposed to replace traditional algorithms. These networks are typically
trained with input vs. synthetic corruption image pairs, with the
goodness of the restoration measured as the network's capability to
correct this synthetic degradation. Plötz and Roh \cite{plotz2017benchmarking}
demonstrated that the synthetic degradation model had significant
limitation, and traditional state-of-the art denoising algorithm BM3D
\cite{burger2012imagedenoising} was still shown to outperform many
deep denoising networks, when the synthetic noise was replaced with
real photographic noise. This highlights the need of creating multiframe
database of multiple modalities from multiple device manufacturers
for realistic evaluation of image restoration networks in general,
as was done by Mayer et al. \cite{mayer2012wavelet} by providing
a freely available multi-frame OCT dataset obtained from \emph{ex
vivo} pig eyes.

\subsubsection{Image restoration}

\label{subsec:Multiframe-reconstruction-techniques}

Most of the literature on multi-frame based deep learning has focused
on super-resolution and denoising. Super-resolution algorithms aim
to improve the spatial resolution of the reconstructed image beyond
what could be obtained from a single input frame. Tao\emph{ et al.}
\cite{tao2017detailrevealing2} implemented a deep learning ``sub-pixel
motion compensation'' network for video input capable of learning
the inter-frame alignment (i.e. image registration) and motion compensation
needed for video super-resolution. In retinal imaging, especially
with OCT the typical problem for efficient super-resolution, are the
retinal motion, lateral resolution limits set by the optical media,
and image noise. Wang \emph{et al. }\cite{wang2018videosuperresolution}
demonstrated using photographic video that motion compensation can
be learned from the data, simplifying dataset acquisition for retinal
deep learning training.

Deblurring (or deconvolution), close to denoising, allows the computational
removal of static and movement blur from acquired images. In most
cases, the exact blurring point-spread-function (PSF) is not known
and has to be estimated (blind deconvolution) from an acquired image
\cite{marrugo2015improving} or sequential images \cite{lian2018deblurring2}.
In retinal imaging, the most common source for image deblurring is
retinal motion \cite{baghaie2017involuntary}, scattering caused by
ocular media opacities \cite{christaras2016intraocular}, and optical
aberrations caused by the optical characteristics of the human eye
itself \cite{burns2018adaptive}. This estimation problem falls under
the umbrella term \emph{inverse problems} that have been solved with
deep learning recently\cite{jin2017deepconvolutional}.

\subsubsection{Physical estimation and correction of the image degradation}

Efficient PSF estimation retinal imaging can be augmented with auxiliary
sensors trying to measure the factors causing retina to move during
acquisition. Retinal vessel pulsations due to pressure fluctuations
during the cardiac cycle can impact the quality. Gating allows imaging
during diastole, when pressure remains almost stable \cite{lee2015cardiacgated}.
Optical methods exist for measuring retinal movement directly using
for example digital micromirror devices (DMD) \cite{vienola2018invivo},
and adaptive optics (AO) systems measuring the dynamic wavefront aberrations
as caused for instance by tear film fluctuations \cite{burns2018adaptive}. 

All these existing physical methods can be combined with deep learning,
providing the measured movements as intermediate targets for the network
to optimize \cite{lee2014deeplysupervised}. Examples of such approaches
are the works by Bollepalli \emph{et al. }\cite{bollepalli2018robustheartbeat}\emph{
}who provided training of the network for robust heartbeat detection
and Li \emph{et al. }\cite{li2018imaging} who have estimated the
blur PSF of light scattered through a glass diffuser simulating the
degradation caused by cataract for retinal imaging. 

Fei \emph{et al. }\cite{fei2017deblurring} used pairs of uncorrected
and adaptive optics-corrected scanning laser ophthalmoscope (AOSLO)
images for learning a\emph{ 'digital adaptive optics'} correction.
This type of adaptive optics -driven network training in practice
might be very useful, providing a cost-effective version of super-resolution
imaging. For example, Jian \emph{et al. }\cite{jian2016lensbased}
proposed to replace deformable mirrors with waveform-correcting lens
lowering the cost and simplifying the optical design \cite{jian2016lensbased},
Carpentras \emph{et al. }\cite{carpentras2017seethrough} demonstrated
a see-through scanning ophthalmoscope without adaptive optics correction,
and very recently a handheld AOSLO imager based on the use of miniature
microelectromechanical systems (MEMS) mirrors was demonstrated by
DuBose \emph{et al.} \cite{dubose2018handheld}.

In practice, all the discussed hardware and software corrections are
not applied simultaneously, i.e. joint image restoration with image
classification \cite{diamond2017dirtypixels}.Thus, the aim of these
operations is to achieve image restoration without loss of clinical
information.

\subsubsection{High-dynamic range (HDR) ophthalmic imaging}

In ophthalmic applications requiring absolute or relative pixel intensity
values for quantitative analysis, as in fundus densitometry \cite{chou2018fundusdensitometry},
or Purkinje imaging for crystalline lens absorption measurements \cite{johnson1997wavelength},
it is desirable to extend the intensity dynamic range from multiple
differently exposed frames using an approach called high dynamic range
(HDR) imaging \cite{zhang2010denoising}. OCT modalities requiring
phase information, such as motion measurement can benefit from higher
bit depths \cite{ling2012theeffects}. Even in simple fundus photography,
the boundaries between optic disc and cup can sometimes be hard to
delineate in some cases due to overexposed optic disc compared to
surrounding tissue, illustrated by \cite{kohler2014multiframe} in
their multiframe reconstruction pipeline. Recent feasibility study
by Ittarat et al. \cite{ittarat2017capability}, showed that HDR acquisition
with tone mapping \cite{zhang2010denoising} of fundus images, visualized
on standard displays, increased the sensitivity but reduced specificity
for glaucoma detection in glaucoma experts. In multimodal or multispectral
acquisition, visible light range acquisition can be enhanced by high-intensity
near-infrared (NIR) strobe \cite{yamashita2017rgbnirimaging} if the
visible light spectral bands do not provide sufficient illumination
for motion-free exposure. The vasculature can be imaged clearly with
NIR strobe for estimating the motion blur between successive visible
light frames \cite{hernandez-matas2017firefundus}.

\subsubsection{Customized spectral filter arrays}

Another operation handled by the ISP is demosaicing \cite{xia2018millionpixel}
which involves interpolation of the color channels. Most color RGB
(red-green-blue) cameras, including fundus cameras include sensors
with a filter grid called Bayer array that is composed of a 2x2 pixel
grid with 2 green, 1 blue and 1 red filter. In fundus imaging, the
red channel has very little contrast, and hypothetically custom demosaicing
algorithms for fundus ISPs may allow for better visualization of clinically
relevant ocular structures. Furthermore, the network training could
be supervised by custom illumination based on light-emitting diodes
(LEDs) for pathology-specific imaging. Bartczak \emph{et al.} \cite{bartczak2017spectrally}
showed that with pathology-optimized illumination, the contrast of
diabetic lesions is enhanced by 30-70\% compared to traditional red-free
illumination imaging.

Recently, commercial sensors with more than 3 color channels have
been released, Omnivision (Santa Clara, California, US) OV4682, for
example, replaced 1 green filter of the Bayer array with a near-infrared
(NIR) filter. In practice, one could acquire continuous fundus video
without pupil constriction using just the NIR channel for the video
illumination, and capturing fundus snapshot simultaneously with a
flash of visible light in addition to the NIR. 

The number of spectral bands on the filter array of the sensor was
extended up 32 bands by imec (Leuven, Belgium). This enables snapshot
multispectral fundus imaging for retinal oximetry \cite{li2017snapshot}.
These additional spectral bands or custom illuminants could also be
used to aid the image processing itself before clinical diagnostics
\cite{ruia2017spectral}. For example, segmenting the macular region
becomes easier with a spectral band around blue 460 nm, as the macular
pigment absorbs strongly at that wavelength and appears darker than
its background on this band \cite{kaluzny2017bayerfilter}.

\subsubsection{Depth-resolved fundus photography}

Traditionally, depth-resolved fundus photography has been done via
stereo illumination of the posterior pole that either involves dual
path optics increasing the design complexity, or operator skill to
take a picture with just one camera \cite{myers2018evolution}. There
are alternatives for depth-resolved fundus camera in a compact form
factor, such as plenoptic fundus imaging that was shown to provide
higher degree of stereopsis than traditional stereo fundus photography
using an off-the-shelf Lytro Illum (acquired by Google, Mountain View,
California, USA) consumer light field camera \cite{palmer2018glarefree}.
Plenoptic cameras however, trade spatial resolution for angular resolution,
for example Lytro Illum has over 40 million pixels, but the final
fundus spatial resolution consists of 635 \texttimes{} 433 pixels.
Simpler optical arrangement for depth imaging with no spatial resolution
trade-off is possible with depth-from-focus algorithms \cite{rivenson2017deeplearning}
that can reconstruct depth map from a sequence of images of different
focus distances ($z$-stack). This rapid switching of focus distances
can be achieved in practice for example by using variable-focus liquid
lenses , as demonstrated for retinal OCT imaging by Cua \emph{et al.}
\cite{cua2016retinal}.

\subsubsection{Compressed sensing}

Especially with OCT imaging, and scanning-based imaging techniques
in general, there is a possibility to use compressed sensing to speed
up the acquisition and reduce the data rate \cite{fang2017segmentation}.
Compressed sensing is based on the assumption that the sampled signal
is sparse in some domain, and thus it can be undersampled and reconstructed
to have a matching resolution for the dense grid. Most of the work
on combined compressed sensing and deep learning has been on magnetic
resonance (MRI) brain scans \cite{schlemper2018adeep}. OCT angiography
(OCTA) is a special variant of OCT imaging that acquires volumetric
images of the retinal and choroidal vasculature through motion contrast
imaging. OCTA acquisition is very sensitive to motion, and would benefit
from sparse sampling with optimized scan pattern \cite{ju2018effective}.

\subsubsection{Defining cost functions}

The design of proper cost function used to define suboptimal parts
of an image is not trivial at all. Early retinal processing work by
Köhler \emph{et al.} \cite{kohler2013automatic} used the retinal
vessel contrast as a proxy measure for image quality, which was implemented
later as fast real-time algorithm by Bendaoudi \emph{et al. }\cite{bendaoudi2018flexible}.
Saha \emph{et al. }\cite{saha2018automated} developed a structure-agnostic
data-driven deep learning network for flagging fundus images either
as acceptable for diabetic retinopathy screening, or as to be recaptured.
In practice, however the cost function used for deep learning training
can be defined in multiple ways as reviewed by Zhao \emph{et al.}
\cite{zhao2017lossfunctions}. They compared different loss functions
for image restoration and showed that the most commonly used $\ell_{2}$
norm (squared error, or ridge regression) was clearly outperformed
in terms of perceptual quality by the multi-scale structural similarity
index (MS-SSIM) \cite{wang2003multiscale}. This was shown to improve
even slightly when the authors combined MS-SSIM with $\ell_{1}$ norm
(absolute deviation, lasso regression). One could hypothesize that
a data-driven quality indicator that reflects the diagnostic differentiation
capability of the image accompanied with perceptual quality, would
be optimal particularly for fundus images.

\subsubsection{Physics-based ground truths}

The unrealistic performance of image restoration networks with synthetic
noise, and the lack of proper real noise benchmark datasets are major
limitations at the moment. Plötz and Roh \cite{plotz2017benchmarking}
created their noise benchmark test by varying the ISO setting of the
camera, and taking the lowest ISO setting as the ground truth ``noise-free''
image. In retinal imaging, construction of good quality ground truth
require some special effort. Mayer \emph{et al.} \cite{mayer2012wavelet}
acquired multiple OCT frames of \emph{ex vivo }pig eyes to avoid motion
artifacts between acquisitions for speckle denoising.

\begin{figure}
\textbf{\includegraphics[width=1\columnwidth]{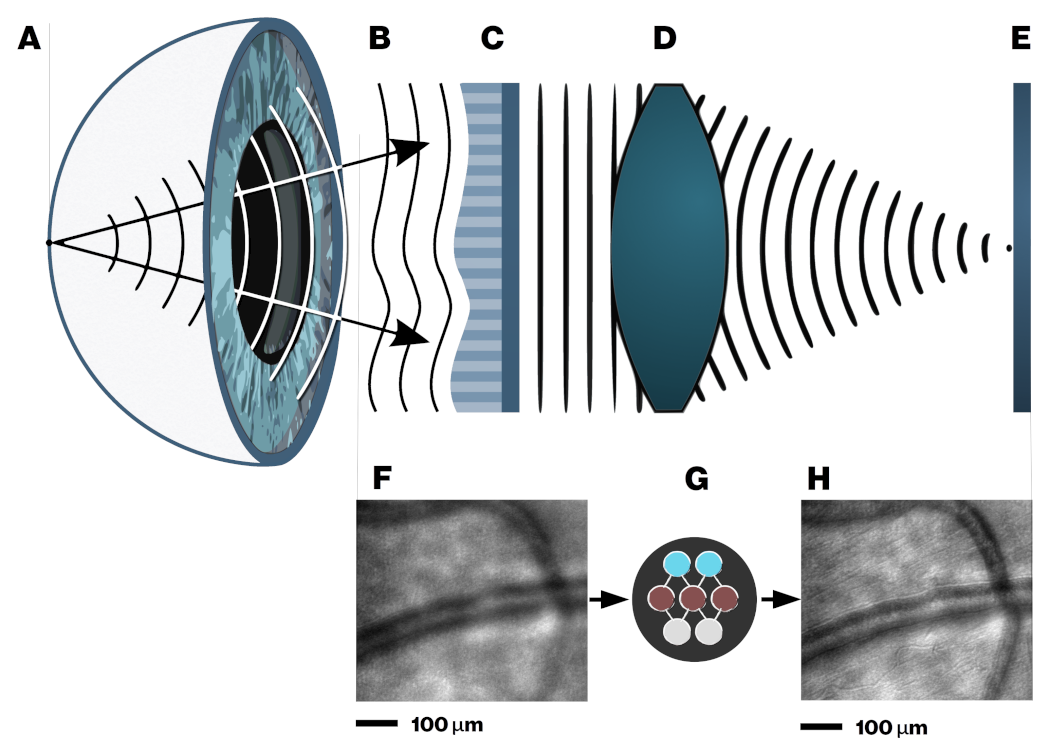}}

\caption{High-level schematic of an adaptive optics retinal imaging system.
The wavefront from retina \textbf{(A) }is distorted mainly by the
cornea and crystalline lens \textbf{(B)}, which is corrected in our
example by lens-based actuator \textbf{(C)} designed for compact imaging
systems \cite{jian2016lensbased}. The imaging optical system \cite{burns2018adaptive}
is illustrated with a single lens for simplicity \textbf{(D)}. The
corrected wavefront on the image sensor \textbf{(E) }is a sharper
version \textbf{(H) }of the image that would be lower quality \textbf{(F)}
without the waveform correction \textbf{(C)}.\emph{ }The ``digital
adaptive optics'' \emph{universal function approximator }\textbf{(G)
}maps the distorted image \textbf{F }to corrected image \textbf{H},
and the network \textbf{G }is the network that was trained with the
image pairs (uncorrected, and corrected). For simplicity, we have
omitted the wavefront sensor from the schematic, and estimate the
distortion in a \emph{sensorless }fashion \cite{burns2018adaptive}.
Images \textbf{F }and \textbf{H }are courtesy of Professor Stephen
A. Burns (School of Optometry, Indiana University) from AOSLO off-axis
illumination scheme for retinal vasculature imaging\cite{chui2012theuse}.
\label{fig:HardwareDL-and-GroundTruth}}
\end{figure}

In humans, commercially available laser speckle reducers can be used
to acquire image pairs with two different levels of speckle noise
\cite{liba2017specklemodulating} (\prettyref{fig:HardwareDL-and-GroundTruth}).
Similar pair for deblurring network training could be acquired with
and without adaptive optics correction \cite{zhang2018aperture} (see
\prettyref{fig:HardwareDL-and-GroundTruth}). In phase-sensitive OCT
application such as elastography, angiography, and vibrometry, a dual
beam setup could be used with a highly phase-stable laser as the ground
truth and ``ordinary'' laser as the input to be enhanced \cite{ling2017highlyphasestable2}. 

Emerging multimodal techniques such as combined OCT and SLO \cite{liu2018transretinal},
and OCT with photoacoustic microscopy (PAM), optical Doppler tomography
(ODT) \cite{leitgeb2014doppler}, and fluorescence microscopy \cite{dadkhah2018amultimodal},
enable interesting joint training from complimentary modalities with
each of them having different strengths. For example, in practice
the lower quality but inexpensive modality could be computationally
enhanced \cite{emami2018generating}

Inter-vendor differences could be further addressed by repeating each
measurement with different OCT machines as taken into account with
clinical diagnosis network by De Fauw\emph{ et al.} \cite{fauw2018clinically2}.
All these hardware-driven signal restorations could be further combined
with existing traditional filters, and use the filter output as targets
for so-called ``copycat'' filters that can estimate existing filters
\cite{gharbi2017deepbilateral}.

\subsubsection{Quantifying uncertainty}

Within the automatic ``active acquisition'' scheme, it is important
to be able to localize the quality problems in an image or in a volume
\cite{kendall2017whatuncertainties}. Leibig \emph{et al. }\cite{leibig2017leveraging}
investigated the commonly used Monte Carlo dropout method \cite{kendall2017whatuncertainties}
for estimating the uncertainty in fundus images for diabetic retinopathy
screening, and its effect on clinical referral decision quality. The
Monte Carlo dropout method improved the identification of substandard
images that were either unusable or had large uncertainty on the model
classification boundaries. Such an approach should, allow rapid identification
of patients with suboptimal fundus images for further clinical evaluation
by an ophthalmologist. 

Similar approach was taken per-patch uncertainty estimation in 3D
super-resolution \cite{tanno2017bayesian}, and in voxel-wise segmentation
uncertainty \cite{eaton-rosen2018towards}. Cobb et al. \cite{cobb2018losscalibrated}
demonstrated an interesting extension to this termed ``loss-calibrated
approximate inference'', that allowed the incorporation of \emph{utility
function} to the network. This utility function was used to model
the asymmetric clinical implications between prediction of \emph{false
negative} and \emph{false positive.}

The financial and quality-of-life cost of an uncertain patch in an
image leading to \emph{false negative }decision might be a lot larger
than \emph{false positive }that might just lead to an additional checkup
by an ophthalmologist.The same utility function could be expanded
to cover disease prevalence \cite{yuan2015thresholdfree}, enabling
end-to-end screening performance to be modeled for diseases such as
glaucoma with low prevalence need very high performance in order to
be cost-efficient to screen \cite{boodhna2016morefrequent}.

\begin{figure}
\textbf{\includegraphics[width=1\columnwidth]{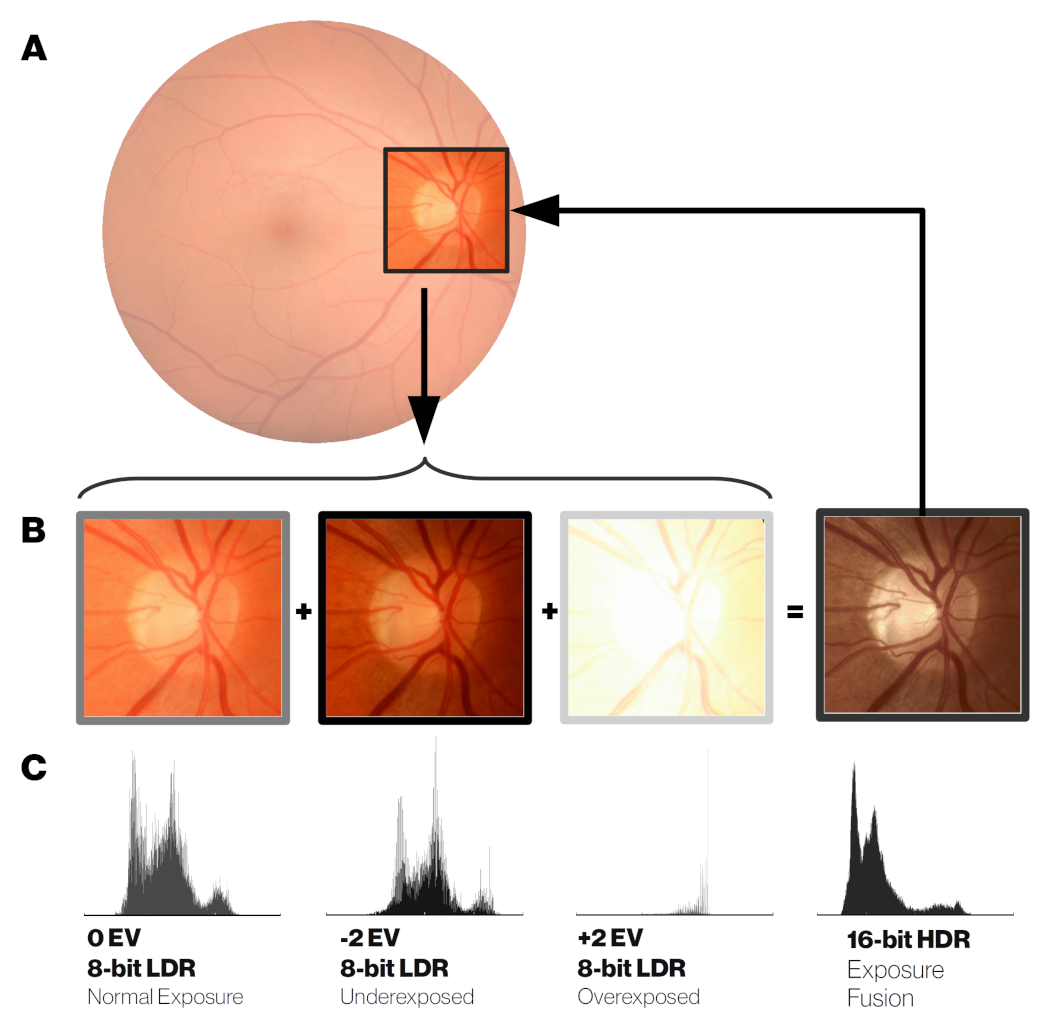}}

\caption{\textbf{(A) }Example of re-acquisition using a region-of-interest
(ROI) defined from the initial acquisition (the full frame). The ROI
has 9\% of the pixels of the full frame making the ROI acquisition
a lot faster if the image sensor allows ROI-based readout. \textbf{(B)
}Multiframe ROI re-acquisition is illustrated with three low-dynamic
range (8-bit LDR) with simulated low-quality camera intensity compression.
The underexposed frame \textbf{(B, left) }exposes optic disc correctly
with less details visible on darker regions of the image as illustrated
by the clipped dark values in histogram \textbf{(C, left, }\emph{clipped
values at }\textbf{0)}, whereas the overexposed frame \textbf{(C,
right) }exposes dark vasculature with detail while overexposing \textbf{(C,
right, }\emph{clipped values at 255}\textbf{)} the bright regions
such as the optic disc. The normal exposure frame \textbf{(B, center)
}is a compromise \textbf{(C, center)} between these two extreme exposures.
\textbf{(D) }When the three LDR frames are combined together using
a exposure fusion technique \cite{li2018multiexposure} into a high-dynamic
range (HDR) image, all the relevant clinical features are exposed
correct possibly improving diagnostics \cite{ittarat2017capability}.
\label{fig:ROI-illustration}}
\end{figure}

The regional uncertainty can then be exploited during active acquisition
by guiding the acquisition iteration to only that area containing
the uncertainly. For example, some CMOS sensors (e.g. Sony IMX250)
allow readout from only a part of the image, faster than one could
do for the full frame. One scenario for smarter fundus imaging could
for example involve initial imaging with the whole field-of-view (FOV)
of the device, followed by multiframe acquisition of only the optic
disc area to ensure that the cup and disc are well distinguishable.,
and that the depth information is of good quality (\prettyref{fig:ROI-illustration}).
Similar active acquisition paradigm is in use for example in drone-based
operator-free photogrammetry. In that application, the drone can autonomously
reconstruct a 3D building model from multiple views recognizing''
where it has not scanned yet, and fly to that location to scan more
\cite{hepp2017plan3dviewpoint}. 

\section{Distributing the computational load}

In typical post-acquisition disease classification studies with deep
learning \cite{ting2017development}, the network training has been
done on large GPU clusters either locally or using cloud-based GPU
servers. However, when embedding deep learning within devices, different
design trade-offs need to be taken into account. Both in hospital
and remote healthcare settings, proper internet connection might be
lacking due to technical infrastructure or institutional policy limitations.
Often the latency requirements are very different for real-time processing
of signals making the use of cloud services impossible \cite{chen2018edgecognitive}.
For example, a lag due to poor internet connection is unacceptable
at intensive care units (ICUs) as those seconds can affect human lives,
and the computing hardware needs to placed next to the sensing device
\cite{davoudi2018theintelligent}.

\subsection{Edge computing}

\begin{figure*}
\textbf{\includegraphics[width=2\columnwidth]{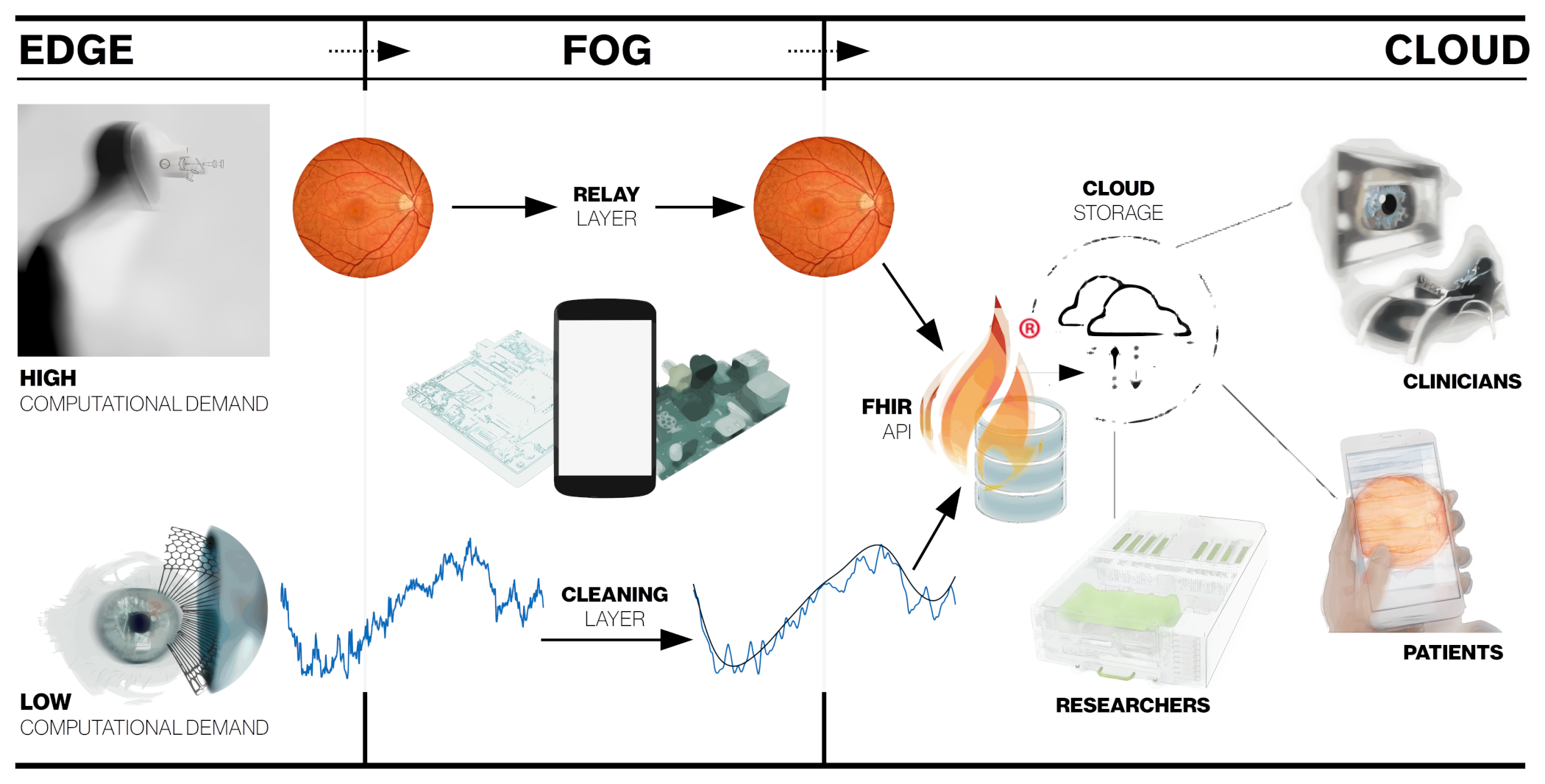}}

\caption{Separation of computations to three different layers. \emph{1) Edge
layer}, which refers to the computations done at the device-level
which in active acquisition ocular imaging (top) require significant
computational power, for example in the form of an embedded GPU. With
wearable intraocular measurement, the contact lens can house only
a very low-power microcontroller (MCU), and it needs to let the \emph{2)
Fog layer }to handle most of the signal cleaning, whereas for ocular
imaging, the fog device mainly just relays the acquired image to 3)\emph{
Cloud layer. }The standardization of the data structure is ensured
through FHIR (Fast Healthcare Interoperability Resources) API (application
programming interface) \cite{mandel2016smarton} before being stored
on secure cloud server. This imaging data along with other clinical
information can then be accessed via healthcare professionals, patients,
and research community. \label{fig:cloud-edge-fog}}
\end{figure*}

In recent years, the concept of\emph{ edge computing} (\prettyref{fig:cloud-edge-fog}A)
has emerged as a complementary or alternative to the cloud computing,
in which computations are done centrally, i.e. away from the ``edge''
. The main driving factor for edge computing are the various Internet-of-Things
(IoT) applications \cite{li2018learning}, or Internet of Medical
Things (IoMT) \cite{chang2018guesteditorial}. Gartner analyst Thomas
Bittman has predicted that the market for processing at the edge,
will expand to similar or increased levels than the current cloud
processing \cite{bittman2017theedge}. Another market research study
by Grand View Research, Inc. \cite{grandviewresearchinc2018edgecomputing},
projected edge computing segment for healthcare \& life sciences to
exceed USD 326 million by 2025. Specifically, the edge computing is
seen as the key enabler of wearables to become a reliable tool for
long-term health monitoring \cite{wang2017areview,nationalinstitutesofhealthnih2018allof}.

\subsection{Fog Computing}

In many cases, an intermediate layer called \emph{fog }or \emph{mist
computing layer} (\prettyref{fig:cloud-edge-fog}B) is introduced
between the edge device and the cloud layer to distribute the computing
load \cite{barik2018leveraging,farahani2018towards,yousefpour2018allone}.
At simplest level, this 3-layer architecture could constitute of simple
low-power IoT sensor (\emph{edge device}) with some computing power
\cite{szydlo2018enabling}. This IoT device could be for example an
inertial measurement unit (IMU)-based actigraph that sends data real-time
to user's smartphone (\emph{fog }device) which contains more computing
power\emph{ }than the \emph{edge} device for gesture recognition \cite{nweke2018deeplearning}.
The gesture recognition model could be used to detect the falls in
elderly, or send corrective feedback back to \emph{edge }device which
could also contain some actuator or a display. An example of such
actuator could be a tactile buzzer for neurorehabilitation applications\cite{yang2018aniotenabled},
or a motorized stage for aligning a fundus camera relative to the
patient's eye \cite{sumi2018nextgeneration}. The smartphone subsequently
sends the relevant data to the cloud for analyzing long-term patterns
at both individual and population-level \cite{aggarwal2017comorbidity,roesch2017automated}.
Alternatively the sensor itself could do some data cleaning, and have
the fog node to handle the sensor fusion of typical clinical 1D biosignal.
An illustration of this concept is the fusion of depth and thermal
cameras for hand-hygiene monitoring \cite{yeung2018bedside}, including
indoor position tracking sensors to monitor healthcare processes at
a hospital level.

\subsection{Balancing edge and fog computations}

For the hardware used in each node, multiple options exist, and in
the literature very heterogeneous architectures are described for
the whole system\cite{dubey2017fogcomputing,farahani2018towards}.
For example, in the SocialEyes project \cite{hansen2016socialeyes},
the diagnostic tests of MARVIN (for mobile autonomous retinal evaluation)
are implemented on GPU-powered Android tablet (NVIDIA SHIELD). In
their rural visual testing application, the device needs to be transportable
and adapted to the limited infrastructure. In this scenario, most
of the computations are already done at the tablet level, and the
\emph{fog }device could for example be a low-cost community smartphone
/ WIFI link. The data can then be submitted to the cloud holding the
centralized electronic health records \cite{raut2017designand}. If
the local computations required are not very heavy, both the edge
and fog functionalities could be combined into one low-cost Raspberry
Pi board computer \cite{sahu2018applylightweight}. In hospital settings
with large patient volumes, it would be preferable to explore different
task-specific data compression algorithms at the cloud-level to reduce
storage and bandwidth requirements. In a teleophthalmology setting,
the compression could be done already at the edge\textendash level
before cloud transmission \cite{rippel2017realtime}.

In the case of fundus imaging, most of that real-time optimization
would be happening at the device-level, with multiple different hardware
acceleration options \cite{hajirassouliha2018suitability,fey2018special}.
One could rely on a low-cost computer such as Raspberry Pi \cite{pagnutti2017layingthe}
and allow for limited computations \cite{shen2017aportable}. This
can be extended if additional computation power is provided at the
cloud level. In many embedded medical applications, GPU options such
as the NVIDIA's Tegra/Jetson platform \cite{perez2018energyaware},
have been increasingly used. The embedded GPU platforms in practice
offer a good compromise between ease-of-use and computational power
of Raspberry Pi and desktop GPUs, respectively.

In some cases the general-purpose GPU (GPGPU) option might not be
able to provide the energy efficiency needed for the required computation
performance. In this case, field-programmable gate arrays (FPGAs)
\cite{zhao2018towards} may be used as an alternative to embedded
GPU, as demonstrated for retinal image analysis \cite{bendaoudi2017flexible},
and real-time video restoration \cite{hung2018videorestoration}.
FPGA implementation may however be problematic, due to increased implementation
complexity. Custom-designed accelerator chips \cite{kulkarni2018anenergyefficient}
and Application-Specific Integrated Circuits (ASIC) \cite{jouppi2017indatacenter}
offer even higher performance but at even higher implementation complexity.

In ophthalmology, there are only a limited number of wearable devices,
allowing for continuous data acquisition. Although the continuous
assessment of intraocular pressure (IOP) is difficult to achieve,
or even controversial \cite{vitish-sharma2018canthe}, commercial
products by Triggerfish® (Sensimed AG, Switzerland) and EYEMATE® (Implandata
Ophthalmic Products GmbH, Germany) have been cleared by the FDA for
clinical use.

Interesting future direction for these monitoring platform is an integrated
MEMS/microfluidics system \cite{araci2014animplantable} that could
simultaneously monitor the IOP and have a passive artificial drainage
system for the treatment of glaucoma \cite{molaei2018upcoming}. The
continuous IOP measurement could be integrated with ``point structure+function
measures'' for individualized deep learning -driven management of
glaucoma as suggested for the management of age-related macular degeneration
(AMD) \cite{schmidt-erfurth2018machine}.

In addition to pure computational restraints, the size and the general
acceptability of the device by the patients can represent a limiting
factor, requiring a more patient-friendly approach. For example, devices
analyzing eye movements \cite{najjar2017disrupted,asfaw2018doesglaucoma}
or pupillary light responses \cite{najjar2018pupillary} can be better
accepted and implemented when using more practical portable devices
rather than bulky research-lab systems. For example \emph{Zhu et al.}
\cite{zhu2018amultimode} have designed an embedded hardware accelerator
for deep learning inference from image sensors of the augmented/mixed
reality (AR/MR) glasses. 

This could be in future integrated with MEMS-based camera-free eye
tracker chip developed by University of Waterloo spin-off company
AdHawk Microsystems (Kitchener, Ontario, Canada) \cite{sarkar2018systemand}
for functional diagnostics or to quantify retinal motion. In this
example of eye movement diagnostics, most of the computation might
be performed at the device level (\emph{edge}), but the patient could
carry a smartphone or a dedicated Raspberry Pi for further post-processing
and/or transmission to cloud services.

\subsection{Cloud computing}

The \emph{cloud layer }(\prettyref{fig:cloud-edge-fog}C) is used
for centralized data storage, allowing both the healthcare professional
and patients to access the electronic health records for example via
the FHIR (Fast Healthcare Interoperability Resources) API (application
programming interface) \cite{mandel2016smarton}. Research groups
can analyze the records as already demonstrated for deep learning
for retinopathy diagnosis \cite{ting2017development,fauw2018clinically2}.
Detailed analysis of different technical options in the \emph{cloud
layer }is beyond the scope of this article, and interested readers
are referred to the following clinically relevant reviews \cite{ping2018biomedical,muhammed2018ubehealth}.

\section{Discussion}

Here we have reviewed the possible applications of deep learning,
introduced at the ophthalmic imaging device level. This extends well-known
application of deep learning for clinical diagnostics \cite{abr`amoff2018pivotal,ting2017development,fauw2018clinically2}.
Such an ``active acquisition'' aims for automatic optimization of
imaging parameters, resulting in improved image quality, and reduced
variability\cite{lee2017machine}. This active approach can be added
to the existing hardware, or can be combined with novel hardware designs.

The main aim of an embedded intelligent deep learning system, is
to favor acquisition of a high-quality image or recording, without
the intervention of a highly skilled operator, in various environments.
There are various healthcare delivery models, in which embedded deep
learning could be used in future routine eye examination: 1) patients
could self-screen themselves, using a shared device located either
in a community clinic, or at the supermarket, requiring no human supervision,
2) the patients could be imaged by a technician either in a 'virtual
clinic', \cite{kotecha2017atechniciandelivered2}, in a hospital waiting
room before an ophthalmologist appointment, or at the optician\footnote{\href{https://www.aop.org.uk/ot/industry/high-street/2017/05/22/oct-rollout-in-every-specsavers-announced}{https://www.aop.org.uk/ot/industry/high-street/2017/05/22/oct-rollout-in-every-specsavers-announced}},
3) patients could be scanned in remote areas by a mobile general healthcare
practitioner\cite{caffery2017modelsof}, and 4) the patients themselves
could do continuous home monitoring for disease progression \cite{roesch2017automated,hong2018rdpdrich}.
Most of the fundus camera and OCT devices come already with some quality
metrics probing the operator to re-take the image, but so far no commercial
device is offering sufficient automatic reconstruction for examples
in presence of ocular media opacities and/or poorly compliant patients. 

Healthcare systems experiencing shortage of manpower may benefit from
modern automated imaging. Putting more intelligence at the device-level
will relieve the healthcare professionals from clerical care for actual
patient care \cite{verghese2018howtech}. With the increased use of
artificial intelligence, the role of the clinician will evolve from
the medical paternalism of the 19th century and evidence-based medicine
of the 20th century, to (big) data-driven clinician working more closely
with intelligent machines and the patients \cite{lerner2018revolution}.
The practical-level interaction with artificial intelligence is not
just near-future science fiction, but very much a reality as the recent
paper on ``augmented intelligence'' in radiology demonstrated \cite{rosenberg2018artificial}.
A synergy between clinicians and AI system resulted in improved diagnostic
accuracy, compared to clinicians' and was better than AI system's
own performance.

At healthcare systems level,intelligent data acquisition will provide
an additional automated data quality verification, resulting in improved
management of data volumes. This is required because size of data
is reported to double every 12-14 months \cite{kilkenny2018dataquality},
addressing, the ``\emph{garbage in - garbage out}``\emph{ }problem
\cite{kilkenny2018dataquality,feldman2018amethodology}. Improved
data quality will also allow more efficient Electronic Health Record
(EHR) mining \cite{shickel2018deepehr}, enabling the healthcare systems
to get closer to the long-term goal of learning healthcare systems
\cite{eisenberg2018shifting} leveraging on prior clinical experience
in structured data/evidence-based sense along with expert clinical
knowledge \cite{lerner2018revolution,thornton2006tacitknowledge}.

Despite the recent developments of deep learning in ophthalmology,
very few prospective clinical trials \emph{per se }have evaluated
its performance in real, everyday life situations. IDx-DR has recently
been approved as the first fully autonomous AI-based FDA-approved
diagnostic system for diabetic retinopathy \cite{abr`amoff2018pivotal},
but the direct benefit of patients, in terms of visual outcome, is
still unclear \cite{keane2018withan} Future innovations emerging
from tech startups, academia, or from established companies will hopefully
improve the quality of the data, through cross-disciplinary collaboration
of designers, engineers and clinicians \cite{depasse2014lessnoise,borsci2018designing},
resulting in improved outcomes of patients with ophthalmic conditions.

\section*{Acknowledgements}

National Health Innovation Centre Singapore Innovation to Develop
(I2D) Grant (NHIC I2D) {[}NHIC-I2D-1708181{]}. We would like to acknowledge
Professor Stephen Burns (Indiana University) for providing images
to illustrate the adaptive optics deep learning correction.

\section*{Disclosures}

The authors declare that there are no conflicts of interest related
to this article.

\end{document}